# The *Do*-Calculus Revisited
# Judea Pearl
# Keynote Lecture, August 17, 2012
# UAI-2012 Conference, Catalina, CA

## Abstract

The *do*-calculus was developed in 1995 to facilitate the identification of causal effects in non-parametric models. The completeness proofs of [Huang and Valtorta, 2006] and [Shpitser and Pearl, 2006] and the graphical criteria of [Tian and Shpitser, 2010] have laid this identification problem to rest. Recent explorations unveil the usefulness of the *do*-calculus in three additional areas: mediation analysis [Pearl, 2012], transportability [Pearl and Bareinboim, 2011] and meta-synthesis. Meta-synthesis (freshly coined) is the task of fusing empirical results from several diverse studies, conducted on heterogeneous populations and under different conditions, so as to synthesize an estimate of a causal relation in some target environment, potentially different from those under study. The talk surveys these results with emphasis on the challenges posed by meta-synthesis. For background material, see ⟨http://bayes.cs.ucla.edu/csl_papers.html⟩.

## 1 Introduction

Assuming readers are familiar with the basics of graphical models, I will start by reviewing the problem of nonparametric identification and how it was solved by the *do*-calculus and its derivatives. I will then show how the *do*-calculus benefits mediation analysis (Section 2), transportability problems (Section 3) and meta-synthesis (Section 4).

### 1.1 Causal Models, interventions, and Identification

**Definition 1.** *(Structural Equation Model) [Pearl, 2000, p. 203].*
*A structural equation model (SEM) $M$ is defined as follows:*

1. *A set $U$ of background or exogenous variables, representing factors outside the model, which nevertheless affect relationship within the model.*

2. *A set $V = \{V_1, \ldots, V_n\}$ of observed endogenous variables, where each $V_i$ is functionally dependent on a subset $PA_i$ of $U \cup V \setminus \{V_i\}$.*

3. *A set $F$ of functions $\{f_1, \ldots, f_n\}$ such that each $f_i$ determines the value of $V_i \in V, v_i = f_i(pa_i, u)$.*

4. *A joint probability distribution $P(u)$ over $U$.*

When an instantiation $U = u$ is given, together with $F$, the model is said to be "completely specified" (at the unit level) when the pair $\langle P(u), F \rangle$ is given, the model is "fully specified" (at the population level). Each fully specified model defines a *causal diagram* $G$ in which an arrow is drawn towards $V_i$ from each member of its parent set $PA_i$.

Interventions and counterfactuals are defined through a mathematical operator called $do(x)$, which simulates physical interventions by deleting certain functions from the model, replacing them with a constant $X = x$, while keeping the rest of the model unchanged. The resulting model is denoted $M_x$.

The postintervention distribution resulting from the action $do(X = x)$ is given by the equation

$$P_M(y|do(x)) = P_{M_x}(y) \qquad (1)$$

In words, in the framework of model $M$, the postintervention distribution of outcome $Y$ is defined as the probability that model $M_x$ assigns to each outcome level $Y = y$. From this distribution, which is readily computed from any fully specified model $M$, we are able to assess treatment efficacy by comparing aspects of this distribution at different levels of $x$. Counterfactuals are defined similarly through the equation $Y_x(u) = Y_{M_x}(u)$ (see [Pearl, 2009, Ch. 7]).

The following definition captures the requirement that a causal query $Q$ be estimable from the data:

**Definition 2.** *(Identifiability) [Pearl, 2000, p. 77].*
A causal query $Q(M)$ is identifiable, given a set of assumptions $A$, if for any two (fully specified) models $M_1$ and $M_2$ that satisfy $A$, we have

$$P(M_1) = P(M_2) \Rightarrow Q(M_1) = Q(M_2) \qquad (2)$$

In words, the functional details of $M_1$ and $M_2$ do not matter; what matters is that the assumptions in $A$ (e.g., those encoded in the diagram) would constrain the variability of those details in such a way that equality of $P$s would entail equality of $Q$s. When this happens, $Q$ depends on $P$ only, and should therefore be expressible in terms of the parameters of $P$.

### 1.2 The Rules of *do*-calculus

When a query $Q$ is given in the form of a *do*-expression, for example $Q = P(y|do(x), z)$, its identifiability can be decided systematically using an algebraic procedure known as the *do*-calculus [Pearl, 1995]. It consists of three inference rules that permit us to map interventional and observational distributions whenever certain conditions hold in the causal diagram $G$.

Let $X, Y, Z$, and $W$ be arbitrary disjoint sets of nodes in a causal DAG $G$. We denote by $G_{\overline{X}}$ the graph obtained by deleting from $G$ all arrows pointing to nodes in $X$. Likewise, we denote by $G_{\underline{X}}$ the graph obtained by deleting from $G$ all arrows emerging from nodes in $X$. To represent the deletion of both incoming and outgoing arrows, we use the notation $G_{\overline{X}\underline{Z}}$.

The following three rules are valid for every interventional distribution compatible with $G$.

**Rule 1** (Insertion/deletion of observations):

$$P(y|do(x), z, w) = P(y|do(x), w)$$
$$\text{if } (Y \perp\!\!\!\perp Z | X, W)_{G_{\overline{X}}} \qquad (3)$$

**Rule 2** (Action/observation exchange):

$$P(y|do(x), do(z), w) = P(y|do(x), z, w)$$
$$\text{if } (Y \perp\!\!\!\perp Z | X, W)_{G_{\overline{X}\underline{Z}}} \qquad (4)$$

**Rule 3** (Insertion/deletion of actions):

$$P(y|do(x), do(z), w) = P(y|do(x), w)$$
$$\text{if } (Y \perp\!\!\!\perp Z | X, W)_{\overline{X Z(W)}}, \qquad (5)$$

where $Z(W)$ is the set of $Z$-nodes that are not ancestors of any $W$-node in $G_{\overline{X}}$.

To establish identifiability of a query $Q$, one needs to repeatedly apply the rules of *do*-calculus to $Q$, until the final expression no longer contains a *do*-operator[1]; this renders it estimable from non-experimental data, and the final *do*-free expression can serve as an *estimator* of $Q$. The *do*-calculus was proven to be complete to the identifiability of causal effects [Shpitser and Pearl, 2006; Huang and Valtorta, 2006], which means that if the *do*-operations cannot be removed by repeated application of these three rules, $Q$ is not identifiable.

Parallel works by [Tian and Pearl, 2002] and [Shpitser and Pearl, 2006] have led to graphical criteria for verifying the identifiability of $Q$ as well as polynomial time algorithms for constructing an estimator of $Q$. This, from a mathematical viewpoint, closes the chapter of nonparametric identification of causal effects.

## 2 Using *do*-Calculus for Identifying Direct and Indirect Effects

Consider the mediation model in Fig. 1(a). (In this section, $M$ stands for the mediating variable.) The *Controlled Direct Effect* ($CDE$) of $X$ on $Y$ is defined as

$$CDE(m) = E(Y|do(X=1), M=m)) - E(Y|do(X=0, M=m))$$

and the *Natural Direct Effect* ($NDE$) is defined by the counterfactual expression

$$NDE = E(Y_{X=1, M_{X=0}}) - E(Y_{X=0})$$

The natural direct effect represents the effect transmitted from $X$ and $Y$ while keeping some intermediate variable $M$ at whatever level it attained prior to the transition [Robins and Greenland, 1992; Pearl, 2001].

Since $CDE$ is a *do*-expression, its identification is fully characterized by the *do*-calculus. The $NDE$, however, is counterfactual and requires more intricate conditions, as shown in Pearl (2001). When translated to graphical language these conditions read:

**Assumption-Set $A$ [Pearl, 2001]**

There exists a set $W$ of measured covariates such that:

A-1 No member of $W$ is a descendant of $X$.

A-2 $W$ blocks all back-door paths from $M$ to $Y$, disregarding the one through $X$.

A-3 The $W$-specific effect of $X$ on $M$ is identifiable using *do*-calculus.

---

[1] Such derivations are illustrated in graphical details in [Pearl, 2009, p. 87].

*A*-4 The $W$-specific joint effect of $\{X, M\}$ on $Y$ is identifiable using *do*-calculus.

The bulk of the literature on mediation analysis has chosen to express identification conditions in the language of "ignorability" (i.e., independence among counterfactuals) which is rather opaque and has led to significant deviation from assumption set $A$.

A typical example of overly stringent conditions that can be found in the literature reads as follows:

> "Imai, Keele and Yamamoto (2010) showed that the sequential ignorability assumption must be satisfied in order to identify the average mediation effects. This key assumption implies that the treatment assignment is essentially random after adjusting for observed pretreatment covariates and that the assignment of mediator values is also essentially random once both observed treatment and the same set of observed pretreatment covariates are adjusted for." [Imai, Jo, and Stuart, 2011]

When translated to graphical representation, these conditions read:

**Assumption-Set $B$**

There exists a set $W$ of measured covariates such that:

*B*-1 No member of $W$ is a descendant of $X$.

*B*-2 $W$ blocks all back-door paths from $X$ to $M$.

*B*-3 $W$ and $X$ block all back-door paths from $M$ to $Y$.

We see that assumption set $A$ relaxes that in $B$ in two ways.

First, we need not insist on using "the same set of observe pretreatment covariates," two separate sets can sometimes accomplish what the same set does not. Second, conditions *A*-3 and *A*-4 invoke *do*-calculus and thus open the door for identification criteria beyond back-door adjustment (as in *B*-2 and *B*-3).

In the sequel we will show that these two features endow $A$ with greater identification power. (See also [Shpitser, 2012].)

### 2.1 Divide and conquer

Fig. 2 demonstrates how the "divide and conquer" flexibility translates into an increase identification power. Here, the $X \to M$ relationship requires an adjustment for $W_2$, and the $X \to Y$ relationship requires an adjustment for $W_3$. If we make the two adjustments separately, we can identify $NDE$ by the estimand:

$$\begin{aligned}
NDE = \sum_m \sum_{w_2, w_3} & P(W_2 = w_2, W_3 = w_3) \\
& [E(Y \mid X = 1, M = m, W_3 = w_3)] \\
& - [E(Y \mid X = 0, M = m, W_3 = w_3,)] \\
& P(M = m \mid X = 0, W_2 = w_2) \quad (6)
\end{aligned}$$

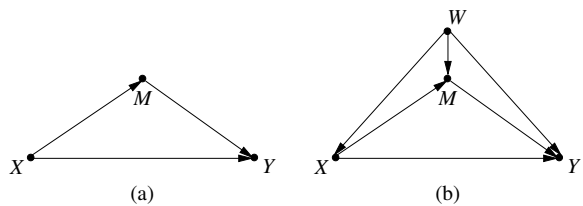

Figure 1: (a) The basic unconfounded mediation model, showing the treatment $(X)$ mediator $(M)$ and outcome $(Y)$. (b) The mediator model with an added covariate $(W)$ that confounds both the $X \to M$ and the $M \to Y$ relationships.

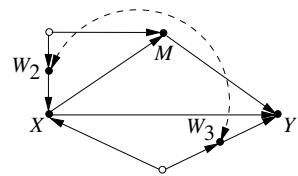

Figure 2: A mediation model with two dependent confounders, permitting the decomposition of Eq. (6). Hollow circles stand for unmeasured confounders. The model satisfies condition $A$ but violates condition $B$.

However, if we insist on adjusting for $W_2$ and $W_3$ simultaneously, as required by assumption set $B$, the $X \to M$ relationship would become confounded, by opening two colliders in tandem along the path $X \to W_3 \leftrightarrow W_2 \leftarrow M$. As a result, assumption set $B$ would deem the $NDE$ to be unidentifiable; there is no covariates set $W$ that simultaneously deconfounds the two relationships.

### 2.2 Going beyond back-door adjustment

Figure 3 displays a model for which the natural direct effect achieves its identifiability through multi-step adjustment (in this case using the front-door procedure), permitted by $A$, though not through a single-step adjustment, as demanded by $B$. In this model, the null set $W = \{0\}$ satisfies conditions *B*-1 and *B*-3, but not condition *B*-2; there is no set of covariates that would enable us to deconfound the treatment-mediator relationship. Fortunately, condition *A*-3 requires only that we identify the effect of $X$ on $M$ by *some do-calculus method*, not necessarily by rendering $X$ random or unconfounded (or ignorable). The presence of observed variable $Z$ permits us to identify this causal effect using the front-door condition [Pearl, 1995; 2009].

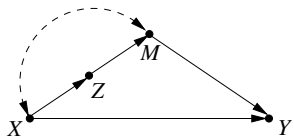

Figure 3: Measuring $Z$ permits the identification of the effect of $X$ on $M$ through the front-door procedure.

In Fig. 4, the front-door estimator needs to be applied to both the $X \to M$ and the $X \to Y$ relationships. In addition, conditioning on $W$ is necessary, in order to satisfy condition $A$-2. Still, the identification of $P(m|do(x), w)$ and $E(Y|do(m,x), w)$ presents no special problems to students of causal inference [Shpitser and Pearl, 2006].

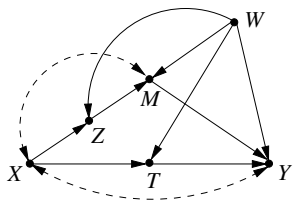

Figure 4: Measuring $Z$ and $T$ permits the identification of the effect of $X$ on $M$ and $X$ on $Y$ for each specific $w$ and leads to the identification of the natural direct effect.

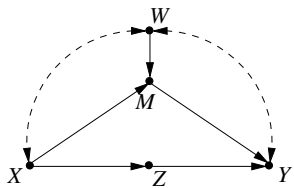

Figure 5: $NDE$ is identified by adjusting for $W$ and using $Z$ to deconfound the $X \to Y$ relationship.

Figure 5, demonstrates the role that an observed covariate $(Z)$ on the $X \to Y$ pathway can play in the identification of natural effects. In this model, conditioning on $W$ deconfounds both the $M \to Y$ and $Y \to M$ relationships but confounds the $X \to Y$ relationship. However the $W$-specific joint effect of $\{X, M\}$ on $Y$ is identifiable through observations on $Z$ (using the front-door estimand).

In Fig. 6, a covariate $Z$ situated along the path from $M$ to $Y$ leads to identifying $NDE$. Here the mediator→outcome relationship is unconfounded (once we fix $X$), so, we are at liberty to choose $W = \{0\}$ to satisfy condition $A$-2. The treatment→ mediator relationship is confounded, and requires an adjustment for $T$ (so does the treatment-outcome relationship). However, conditioning on $T$ will confound the $\{MX\} \to Y$ relationship (in violation of condition $A$-4). Here, the presence of $Z$ comes to our help, for it permits us to estimate $P(Y \mid do(x, m), t)$ thus rendering $NDE$ identifiable.

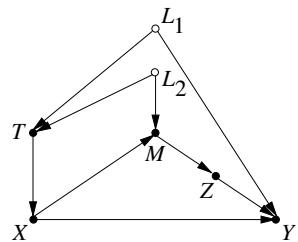

Figure 6: The confounding created by adjusting for $T$ can be removed using measurement of $Z$.

## 3 Using *do*-Calculus to Decide Transportability

In applications involving identifiability, the role of the *do*-calculus is to remove the *do*-operator from the query expression. We now discuss a totally different application, to decide if experimental findings can be transported to a new, potentially different environment, where only passive observations can be performed. This problem, labeled "transportability" in [Pearl and Bareinboim, 2011] can also be reduced to syntactic operation using the *do*-calculus but here the aim will be to separate the *do*-operator from a set $S$ of variables, that indicate disparities between the two environment.

We shall motivate the problem through the following three examples.

**Example 1.** *We conduct a randomized trial in Los Angeles (LA) and estimate the causal effect of exposure $X$ on outcome $Y$ for every age group $Z = z$ as depicted in Fig. 7(a). We now wish to generalize the results to the population of New York City (NYC), but data alert us to the fact that the study distribution $P(x, y, z)$ in LA is significantly different from the one in NYC (call the latter $P^*(x, y, z)$). In particular, we notice that the average age in NYC is significantly higher than that in LA. How are we to estimate the causal effect of $X$ on $Y$ in NYC, denoted $P^*(y|do(x))$.*

**Example 2.** *Let the variable $Z$ in Example 1 stand for subjects language proficiency, and let us assume that*

*Z does not affect exposure (X) or outcome (Y), yet it correlates with both, being a proxy for age which is not measured in either study (see Fig. 7(b)). Given the observed disparity $P(z) \neq P^*(z)$, how are we to estimate the causal effect $P^*(y|do(x))$ for the target population of NYC from the z-specific causal effect $P(y|do(x), z)$ estimated at the study population of LA?*

**Example 3.** *Examine the case where $Z$ is a X-dependent variable, say a disease bio-marker, standing on the causal pathways between $X$ and $Y$ as shown in Fig. 7(c). Assume further that the disparity $P(z) \neq P^*(z)$ is discovered in each level of $X$ and that, again, both the average and the z-specific causal effect $P(y|do(x), z)$ are estimated in the LA experiment, for all levels of $X$ and $Z$. Can we, based on information given, estimate the average (or z-specific) causal effect in the target population of NYC?*

To formalize problems of this sort, Pearl and Bareinboim devised a graphical representation called "selection diagrams" which encodes knowledge about differences between populations. It is defined as follows:

**Definition 3.** *(Selection Diagram).*
*Let $\langle M, M^* \rangle$ be a pair of structural causal models (Definition 1) relative to domains $\langle \Pi, \Pi^* \rangle$, sharing a causal diagram G. $\langle M, M^* \rangle$ is said to induce a selection diagram D if D is constructed as follows:*

1. *Every edge in G is also an edge in D;*

2. *D contains an extra edge $S_i \to V_i$ whenever there exists a discrepancy $f_i \neq f_i^*$ or $P(U_i) \neq P^*(U_i)$ between M and $M^*$.*

In summary, the S-variables locate the mechanisms where structural discrepancies between the two populations are suspected to take place. Alternatively, the absence of a selection node pointing to a variable represents the assumption that the mechanism responsible for assigning value to that variable is the same in the two populations, as shown in Fig. 7.

Using selection diagrams as the basic representational language, and harnessing the concepts of intervention, *do*-calculus, and identifiability (Section 1), we can now give the notion of transportability a formal definition.

**Definition 4.** *(Transportability)*
*Let D be a selection diagram relative to domains $\langle \Pi, \Pi^* \rangle$. Let $\langle P, I \rangle$ be the pair of observational and interventional distributions of $\Pi$, and $P^*$ be the observational distribution of $\Pi^*$. The causal relation $R(\Pi^*) = P^*(y|do(x), z)$ is said to be transportable from $\Pi$ to $\Pi^*$ in D if $R(\Pi^*)$ is uniquely computable from $P, P^*, I$ in any model that induces D.*

**Theorem 1.** *Let D be the selection diagram characterizing two populations, $\Pi$ and $\Pi^*$, and S a set*

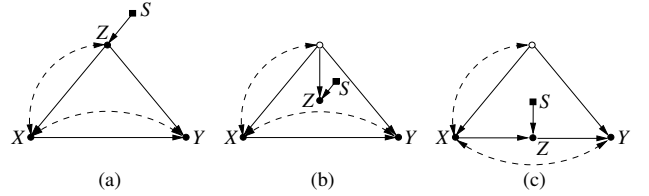

Figure 7: Selection diagrams depicting Examples 1–3. In (a) the two populations differ in age distributions. In (b) the populations differs in how $Z$ depends on age (an unmeasured variable, represented by the hollow circle) and the age distributions are the same. In (c) the populations differ in how $Z$ depends on $X$.

*of selection variables in D. The relation $R = P^*(y|do(x), z)$ is transportable from $\Pi$ to $\Pi^*$ if the expression $P(y|do(x), z, s)$ is reducible, using the rules of do-calculus, to an expression in which S appears only as a conditioning variable in do-free terms.*

This criterion was proven to be both sufficient and necessary for causal effects, namely $R = P(y|do(x))$ [Bareinboim and Pearl, 2012b]. Theorem 1 does not specify the sequence of rules leading to the needed reduction when such a sequence exists. [Bareinboim and Pearl, 2012b] established a complete and effective graphical procedure of confirming transportability which also synthesizes the transport formula whenever possible. For example, the transport formulae derived for the three models in Fig. 7 are (respectively):

$$P^*(y|do(x)) = \sum_z P(y|do(x), z) P^*(z) \qquad (7)$$

$$P^*(y|do(x)) = p(y|do(x)) \qquad (8)$$

$$P^*(y|do(x)) = \sum_z P(y|z, x) P^*(z|x) \qquad (9)$$

Each transport formula determines for the investigator what information need to be taken from the experimental and observational studies and how they ought to be combined to yield an unbiased estimate of $R$.

## 4 From "Meta-analysis" to "Meta-synthesis"

"Meta analysis" is a data fusion problem aimed at combining results from many experimental and observational studies, each conducted on a different population and under a different set of conditions, so as to synthesize an aggregate measure of effect size that is "better," in some sense, than any one study in isolation. This fusion problem has received enormous attention in the health and social sciences, where data are scarce and experiments are costly.

Unfortunately, current techniques of meta-analysis do little more than take weighted averages of the various studies, thus averaging apples and oranges to infer properties of bananas. One should be able to do better. Using "selection diagrams" to encode commonalities among studies, we should be able to "synthesize" an estimator that is guaranteed to provide unbiased estimate of the desired quantity based on information that each study share with the target environment. The basic idea is captured in the following definition and theorem.

**Definition 5.** *(Meta-identifiability):*
*A relation $R$ is said to be "meta-identifiable" from a set of populations $(\Pi_1, \Pi_2, \ldots, \Pi_K)$ to a target population $\Pi^*$ iff it is identifiable from the information set $I = \{I(\Pi_1), I(\Pi_2), \ldots, I(\Pi_K), I(\Pi^*)\}$, where $I(\Pi_k)$ stands for the information provided by population $\Pi_k$.*

**Theorem 2.** *(Meta-identifiability):*
*Given a set of studies $\{\Pi_1, \Pi_2, \ldots, \Pi_K\}$ characterized by selection diagrams $\{D_1, D_2, \ldots, D_K\}$ relative to a target population $\Pi^*$, a relation $R(\Pi^*)$ is "meta-identifiable" if it can be decomposed into a set of sub-relations of the form:*

$$R_k = P(V_k|do(W_k), Z_k) \qquad k = 1, 2, \ldots, K$$

*such that each $R_k$ is transportable from some $D_k$.*

Theorem 2 reduces the problem of Meta synthesis to a set of transportability problems, and calls for a systematic way of decomposing $R$ to fit the information provided by $I$.

**Exemplifying meta-synthesis**

Consider the diagrams depicted in Fig. 8, each representing a study conducted on a different population and under a different set of conditions. Solid circles represent variables that were measured in the respective study and hollow circles variables that remained unmeasured. An arrow ■→ represents an external influence affecting a mechanism by which the study population is assumed to differ from the target population $\Pi^*$, shown in Fig. 8(a). For example, Fig. 8(c) represents an observational study on population $\Pi_c$ in which variables $X, Z$ and $Y$ were measured, $W$ was not measured and the prior probability $P_c(z)$ differs from that of the target population $P^*(z)$. Diagrams (b)–(f) represent observational studies while (g)–(j) stand for experimental studies with $X$ randomized (hence the missing arrows into $X$).

Despite differences in populations, measurements and conditions, each of the studies may provide information that bears on the target relation $R(\Pi^*)$ which, in this example, we take to be the causal effect of $X$ on $Y$, $P^*(y|do(x))$ or; given the structure of Fig. 8(a),

$$R(\Pi^*) = P^*(y|do(x)) = \sum_z P^*(y|x,z)P^*(z).$$

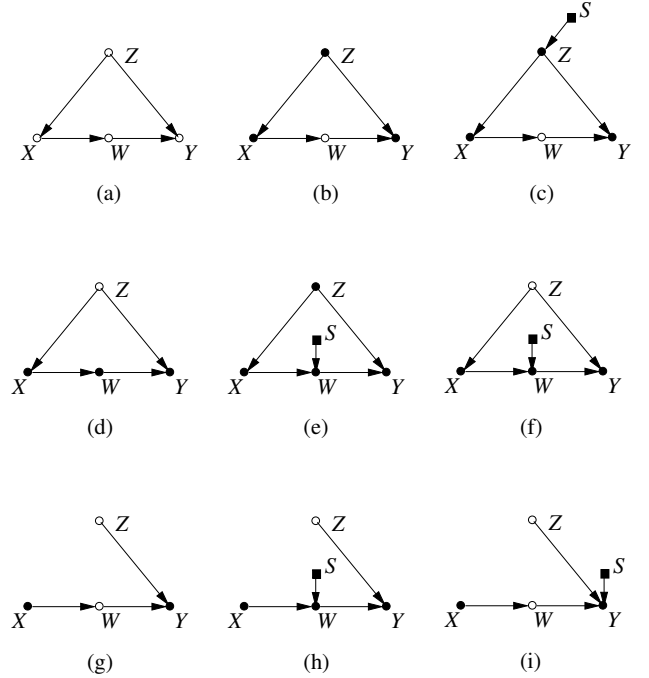

Figure 8: Diagrams representing 8 studies ((b)–(i)) conducted under different conditions on different populations, aiming to estimate the causal effect of $X$ on $Y$ in the target population, shown in 8(a).

While $R(\Pi^*)$ can be estimated directly from some of the studies, (e.g., (g)) or indirectly from others (e.g., (b) and (d)), it cannot be estimated from those studies in which the population differs substantially from $\Pi^*$, (e.g., (c), (e), (f)). The estimates of $R$ provided by the former studies may differ from each other due to sampling variations and measurement errors, and can be aggregated in the standard tradition of meta analysis. The latter studies, however, should not be averaged with the former, since they do not provide unbiased estimates of $R$. Still, they are not totally useless, for they can provide information that renders the former estimates more accurate. For example, although we cannot identify $R$ from study 8(c), since $P_c(z)$ differs from the unknown $P^*(z)$, we can nevertheless use the estimates of $P_c(x|z), P_c(y|z,x)$ that 8(c) provides to improve the accuracy of $P^*(x|z)$ and $P^*(y|z,x)$[2] which may be needed for estimating $R$ by indirect meth-

---

[2]The absence of boxed arrows into $X$ and $Y$ in Fig. 8(c) implies the equalities

$$P_c(x|z) = P^*(x|z) \text{ and } P_c(y|z,x) = P^*(y|z,x).$$

ods. For example, $P^*(y|z,x)$ is needed in study 8(b) if we use the estimator $R = \sum_z P^*(y|x,z)P^*(z)$, while $P^*(x|z)$ is needed if we use the inverse probability estimator $R = \sum_z P^*(x,y,z)/P^*(x|z)$.

Similarly, consider the randomized studies depicted in 8(h) and 8(i). None is sufficient for identifying $R$ in isolation yet, taken together, they permit us to borrow $P_i(w|do(x))$ from 8(i) and $P_h(y|w,do(x))$ from 8(h) and synthesize a bias-free estimator:

$$R = \sum_w P^*(y|w,do(x))P^*(w|do(x))$$
$$= \sum_w P_h(y|w,do(x)P_i(w|do(x))$$

The challenge of meta synthesis is to take a collection of studies, annotated with their respective selection diagrams (as in Fig. 8), and construct an estimator of a specified relation $R(\Pi^*)$ that makes maximum use of the samples available, by exploiting the commonalities among the populations studied and the target population $\Pi^*$. As the relation $R(\Pi^*)$ changes, the synthesis strategy will change as well.

It is hard not to speculate that data-pooling strategies based on the principles outlined here will one day replace the blind methods currently used in meta analysis.

**Knowledge-guided Domain Adaptation**

It is commonly assumed that causal knowledge is necessary only when interventions are contemplated and that in purely predictive tasks, probabilistic knowledge suffices. When dealing with generalization across domains, however, causal knowledge can be valuable, and in fact necessary even in predictive or classification tasks.

The idea is simple; causal knowledge is essentially knowledge about the mechanisms that remain *invariant* under change. Suppose we learn a probability distribution $P(x,y,z)$ in one environment and we ask how this probability would change when we move to a new environment that differs slightly from the former. If we have knowledge of the causal mechanism generating $P$, we could then represent where we suspect the change to occur and channel all our computational resources to re-learn the local relationship that has changed or is likely to have changed, while keeping all other relationships invariant.

For example, assume that $P^*(x,y,z)$ is the probability distribution in the target environment and our interest lies in estimating $P^*(x|z)$ knowing that the causal diagram behind $P$ is the chain $X \rightarrow Y \rightarrow Z$, and that the process determining $Y$, represented by $P^*(y|x)$, is the only one that changed. We can simply re-learn $P^*(y|x)$ and estimate our target relation $P^*(x|z)$ without measuring $Z$ in the new environment. (This is done using $P^*(x,y,z) = P(x)P^*(y|x)P(z|y)$ with the first and third terms transported from the source environment.)

In complex problems, the savings gained by focusing on only a small subset of variables in $P^*$ can be enormous, because any reduction in the number of measured variables translates into substantial reduction in the number of samples needed to achieve a given level of prediction accuracy.

As can be expected, for a given transported relation, the subset of variables that can be ignored in the target environment is not unique, and should therefore be chosen so as to minimize both measurement costs and sampling variability. This opens up a host of new theoretical questions about transportability in causal graphs. For example, deciding if a relation is transportable when we forbid measurement of a given subset of variables in $P^*$ [Pearl and Bareinboim, 2011], or deciding how to pool studies optimally when sample size varies drastically from study to study [Pearl, 2012].

## Conclusions

The *do*-calculus, which originated as a syntactic tool for identification problems, was shown to benefit three new areas of investigation. Its main power lies in reducing to syntactic manipulations complex problems concerning the estimability of causal relations under a variety of conditions.

In Section 2 we demonstrated that going beyond standard adjustment for covariates, and unleashing the full power of *do*-calculus, can lead to improved identification power of natural direct and indirect effects. Section 3 demonstrated how questions of transportability can be reduced to symbolic derivations in the *do*-calculus, yielding graph-based procedures for deciding whether causal effects in the target population can be inferred from experimental findings in the study population. When the answer is affirmative, the procedures further identify what experimental and observational findings need be obtained from the two populations, and how they can be combined to ensure bias-free transport.

In a related problem, Bareinboim and Pearl (2012b) show how the *do*-calculus can be used to decide whether the effect of $X$ on $Y$ can be estimated from experiments on a different set, $Z$, that is more accessible to manipulations. Here the aim is to transform *do*-expressions to sentences that invoke only $do(z)$ symbols.

Finally, in Section 4, we tackled the problem of data fusion and showed that principled fusion (which we call meta-synthesis) can be reduced to a sequence of syntactic operations each involving a local transportability exercise. This task leaves many questions unsettled, because of the multiple ways a give relation can be decomposed.

# Acknowledgments

This research was supported in parts by grants from NIH #1R01 LM009961-01, NSF #IIS-1018922, and ONR #N000-14-09-1-0665 and #N00014-10-0933.